\documentclass{article}

\usepackage[utf8]{inputenc}     % UTF-8 인코딩 지원
\usepackage[T1]{fontenc}        % 8-bit 폰트 인코딩
\usepackage{times}              % Times New Roman 폰트
\usepackage{latexsym}           % LaTeX 심볼
\usepackage{inconsolata}        % 고정폭 폰트
\usepackage{microtype}          % 마이크로타이포그래피 개선
\usepackage{amsfonts}           % 수학 기호
\usepackage{amsmath}            % 수학 명령어 (\tfrac 등)
\usepackage{kotex}              % 한글 지원

\usepackage{arxiv}              % arXiv 스타일
\usepackage{hyperref}           % 하이퍼링크
\usepackage{pgfplotstable}      % Table-based plotting
\usepackage{url}                % URL 서식
\usepackage{doi}                % DOI 링크
\usepackage{natbib}             % 참고문헌 스타일
\usepackage{pgf-pie}
\usepackage{booktabs}           % 전문적인 표 디자인
\usepackage{array}              % 향상된 표 기능
\usepackage{tabularx}           % 확장된 표 기능
\usepackage{tabularray}         % 현대적 표 기능
\usepackage{multirow}           % 행 병합
\usepackage{multicol}           % 다중 열
\usepackage{makecell}           % 셀 서식
\usepackage{colortbl}           % 표 색상
\usepackage{hhline}             % 향상된 가로선

\usepackage{graphicx}           % 이미지 포함
\usepackage{subfloat}           % 하위 그림
\usepackage{subcaption}         % 하위 캡션
\usepackage{caption}            % 캡션 커스터마이징
\usepackage{xcolor}             % 색상 지원

\usepackage{enumitem}           % 리스트 커스터마이징
\usepackage{nicefrac}           % 분수 표현
\usepackage{lipsum}             % 더미 텍스트 (테스트용)
\usepackage{changepage}         % 페이지 레이아웃 조정
\usepackage{tcolorbox}          % 색상 박스
\usepackage{ragged2e}           % 정렬 제어
\usepackage{pgfplots}
\usepackage{xcolor}
\usepackage{pgfplots}
\usepackage{tikz}
\usepackage{pgf-pie} % 파이 차트를 위한 패키지
\usepackage{subcaption}
\pgfplotsset{compat=1.18}
\usepackage{xcolor}
\usepackage{pgfplots}
\usepgfplotslibrary{polar}
\pgfplotsset{compat=newest}
\renewcommand{\undertitle}{}

\title{KoBALT: Korean Benchmark for Advanced Linguistic Tasks}

\date{} 					% Or removing it

\author{Hyopil Shin \\
  Seoul National University \\
  \texttt{hpshin@snu.ac.kr} \\
  \And
  Sangah Lee \\
  Seoul National University \\
  \texttt{sanalee@snu.ac.kr} \\
  \And
  Dongjun Jang \\
  Seoul National University \\
  \texttt{qwer4107@snu.ac.kr} \\
  \And
  Wooseok Song \\
  Seoul National University \\
  \texttt{semojak@snu.ac.kr} \\
  \And
  Jaeyoon Kim \\
  Seoul National University \\
  \texttt{toscour345@snu.ac.kr} \\
  \And
  Chaeyoung Oh \\
  Seoul National University \\
  \texttt{nyong10@snu.ac.kr} \\
  \And
  Hyemi Jo \\
  Seoul National University \\
  \texttt{huimei6361@snu.ac.kr} \\
  \And
  Youngchae Ahn \\
  Seoul National University \\
  \texttt{estelle1026@snu.ac.kr} \\
  \And
  Sihyun Oh \\
  Seoul National University \\
  \texttt{osthepublic@snu.ac.kr} \\
  \And
  Hyohyeong Chang \\
  Seoul National University \\
  \texttt{mipiw3842@snu.ac.kr} \\
  \And
  Sunkyoung Kim \\
  LG AI Research \\ 
  \texttt{sunkyoung.kim@lgresearch.ai} \\
  \And
  Jinsik Lee \\
  LG AI Research \\
  \texttt{jinsik.lee@lgresearch.ai} \\ 
}
\renewcommand{\shorttitle}{KoBALT: Korean Benchmark for Advanced Linguistic Tasks}
\hypersetup{
pdftitle={A template for the arxiv style},
pdfsubject={q-bio.NC, q-bio.QM},
pdfauthor={David S.~Hippocampus, Elias D.~Striatum},
pdfkeywords={First keyword, Second keyword, More},
}

\begin{document}

\maketitle

\begin{abstract}
We introduce KoBALT (Korean Benchmark for Advanced Linguistic Tasks), a comprehensive linguistically-motivated benchmark comprising 700 multiple-choice questions spanning 24 phenomena across five linguistic domains: syntax, semantics, pragmatics, phonetics/phonology, and morphology. KoBALT is designed to advance the evaluation of large language models (LLMs) in Korean, a morphologically rich language, by addressing the limitations of conventional benchmarks that often lack linguistic depth and typological grounding. It introduces a suite of expert-curated, linguistically motivated questions with minimal n-gram overlap with standard Korean corpora, substantially mitigating the risk of data contamination and allowing a more robust assessment of true language understanding. Our evaluation of 20 contemporary LLMs reveals significant performance disparities, with the highest-performing model achieving 61\% general accuracy but showing substantial variation across linguistic domains - from stronger performance in semantics (66\%) to considerable weaknesses in phonology (31\%) and morphology (36\%). Through human preference evaluation with 95 annotators, we demonstrate a strong correlation between KoBALT scores and human judgments, validating our benchmark's effectiveness as a discriminative measure of Korean language understanding. KoBALT addresses critical gaps in linguistic evaluation for typologically diverse languages and provides a robust framework for assessing genuine linguistic competence in Korean language models.
\end{abstract}

\section{Introduction}
Recent advancements in Large Language Models (LLMs) have been remarkable, particularly following the emergence of increasingly sophisticated LLMs with varying architectures, parameter scales \citep{megatron2020, gpt42023}, and training objectives \citep{deepseek2025, GPT32020}. These recent models demonstrate impressive capabilities across various linguistic tasks, and many models approach or surpass human-level performance on specific benchmarks. However, in terms of multilingual capabilities including the Korean language, there remains a gap in assessing the true linguistic capabilities of them, which are not adequately captured in recent knowledge-intensive and challenging benchmarks based on translated and repurposed materials without sufficient involvement of linguistic knowledge. 

% Despite the emergence of those powerful models, the evaluation benchmark which is required for measuring if a language model truely is competitive lacks of its validity, especially on linguistic competence. Some benchmarks merely focuse the model's general knowledge, which do not regard the models' linguistic ability, although some challenging benchmarks do utilize linguistic exams. The score over those benchmarks does not prove whether the model has high competence as well as its preference to human. 

There are two streams of benchmarks that are linguistically motivated. A traditional stream comprises benchmarks such as BLiMP \citep{blimp2019}, Holmes \citep{holmes2024}, and SyntaxGym \citep{gauthier2020syntaxgym}. However, these benchmarks were built with highly controlled sentences, which fall short in evaluating models' advanced linguistic ability in multiple domains: phonetics, phonology, morphology, syntax, semantics, and pragmatics. Another stream consists of benchmarks sourced from the linguistic olympiad(\cite{linguini2024}, \cite{IOLBENCH2025}, \cite{Bean2024LINGOLYAB}). This type of benchmarks merely focus on evaluating complex deductive reasoning based on linguistic data in a puzzle-like manner, whose results are apart from the model's linguistic competence on multiple levels of individual languages. Moreover, when it comes to Korean language, many of existing benchmarks have additional problems that they are constructed via translations of English benchmarks, which reduces their validity on the model competence.

We introduce KoBALT (Korean Benchmark for Advanced Linguistic Tasks), comprising 700 linguistically motivated multiple-choice questions spanning 24 distinct linguistic phenomena across five fundamental domains: syntax (300), semantics (215), pragmatics (81), phonetics/phonology (62), and morphology (42). Unlike most existing Korean benchmarks, KoBALT features original, linguist-crafted test items with minimal n-gram overlap with common Korean training corpora (bigrams <8.6\%, trigrams <0.7\%), mitigating data contamination concerns. We evaluate 20 contemporary LLMs on KoBALT, including both proprietary (Claude 3.7, GPT-4o) and open-source models (LLaMA, Mistral, Qwen). Our results reveal significant performance disparities, with the highest-performing model (Claude 3.7 Sonnet) achieving 61\% accuracy overall, with notable performance variations across linguistic domains. Models generally demonstrate greater proficiency in semantics (66\%) compared to phonology (31\%) and morphology (36\%).
To ensure KoBALT's validity that our dataset's evaluation results correlate with actual human preferences for certain models, we conduct a human preference evaluation involving 95 annotators. Through Bradley-Terry analysis of their judgments on model responses, we demonstrate a strong correlation between KoBALT scores and human evaluations (r=0.638 for the best-performing model), confirming our benchmark as an effective and discriminative measure of Korean language understanding.

KoBALT is a challenging benchmark designed to elicit and evaluate the deep and sophisticated linguistic capabilities of large language models (LLMs) by integrating linguistic expertise from human experts with the generative capacities of LLMs. By thoroughly reflecting the structural and typological properties of language, KoBALT not only enables fine-grained assessment of Korean, but also serves as a reference framework for constructing linguistically rigorous benchmarks in other languages.

% [Rephrase]
% To validate KoBALT's ecological validity, we conduct a human preference evaluation with 95 annotators providing pairwise judgments on model responses. Through Bradley-Terry analysis of these judgments, we demonstrate a strong correlation between KoBALT scores and human evaluations ($r=0.638$ for the best-performing model), confirming our benchmark's effectiveness as a discriminative measure of Korean language understanding.

\section{Related Work}

\subsection{Benchmarks for Linguistic Evaluation}

% While conventional benchmarks such as GLUE, SuperGLUE, MMLU \citep{MMLU2020}, and BIG-bench \citep{Bigbench2022} test models utilizing either conventional NLP tasks or broad task coverage, their narrow coverage and emphasis on encyclopedic knowledge often limit the depth of linguistic evaluation. This leads to the development of linguistically-controlled benchmarks such as BLiMP \citep{blimp2019}, Holmes \citep{holmes2024}, and more domain-specific evaluations such as SyntaxGym \citep{gauthier2020syntaxgym} for syntax, PUB \citep{pub2024} for pragmatics, and PhonologyBench \citep{Suvarna2024PhonologyBenchEP} for phonology. Some works including Linguini \citep{linguini2024}, IOLBench \citep{IOLBENCH2025}, and LINGOLY \citep{Bean2024LINGOLYAB} conceptualize linguistics for complex reasoning tasks, utilizing linguistics olympiad problems.

% While these tasks pose challenges to models, they tend to assess general reasoning ability rather than linguistic competence. Our work differs from these approaches by introducing problem sets that require both linguistic knowledge and reasoning capabilities addressing specific phenomena of the Korean language. Additionally, we verify our benchmark's validity by demonstrating correlation between benchmark performance and human preference judgments ― a methodological contribution absent in previous linguistic evaluation frameworks.

% [Rephrase]
The evaluation of language models (LMs) has undergone a significant shift from task-oriented benchmarks to linguistically grounded diagnostics. Early frameworks such as GLUE and SuperGLUE \citep{Wang2018GLUEAM,Wang2019SuperGLUEAS} established standard NLP tasks—natural language inference(NLI), question answering, and classification—but were rapidly saturated, limiting their ability to distinguish models with genuine linguistic understanding from those leveraging superficial pattern recognition or dataset-specific cues.

In response, broader-scale benchmarks like MMLU \citep{MMLU2020} and BIG-Bench \citep{Bigbench2022} expanded task coverage across domains. While these benchmarks substantially broadened task coverage, their emphasis on encyclopedic knowledge often limited the depth of linguistic evaluation. These limitations have led to the development of linguistically-targeted resources such as BLiMP \citep{blimp2019}, which focuses on minimal-pair diagnostics for core linguistic phenomena, and Holmes \citep{holmes2024}, which extends this approach by covering a wider range of linguistic subfields. SyntaxGym \citep{gauthier2020syntaxgym} introduced psycholinguistically inspired templates to test syntactic generalization, while domain-specific evaluations such as PUB \citep{pub2024} target core pragmatic phenomena including implicature, presupposition, reference, and deixis. MultiPragEval \citep{multiprag2024} further expands pragmatic evaluation through multilingual tasks based on Grice’s Cooperative Principle, while PhonologyBench \citep{Suvarna2024PhonologyBenchEP} evaluates models’ ability to generalize phonological rules.  

% 문단제거 고민 -> [Delete]
% Beyond these large-scale benchmarks, a number of studies propose focused diagnostic tasks that reveal model limitations in specific linguistic domains. \citet{ismayilzada2024evaluating} construct targeted evaluations to test compositionality in agglutinative languages, \citet{mortensen2024verbing} design tasks for assessing models’ handling of zero-derivation and syntactic flexibility, and \citet{riccardi2024two} examine semantic plausibility through controlled binary judgments. Together, these works contribute to a growing body of linguistically motivated evaluation protocols that move beyond surface performance to probe the structural properties of language understanding in LLMs.

% Another distinct approach includes benchmarks that conceptualize linguistics as a domain for complex reasoning. Works such as Linguini \citep{linguini2024}, IOLBench \citep{IOLBENCH2025}, and LINGOLY \citep{Bean2024LINGOLYAB} leverage problem sets from linguistics olympiads, emphasizing deductive reasoning through tasks like unseen language translation, cognate identification, and graphophonemic transcription. 

% [Rephrase]
A distinct methodological approach has emerged through benchmarks conceptualizing linguistics as a domain for complex reasoning. Works such as Linguini \citep{linguini2024}, IOLBench \citep{IOLBENCH2025}, and LINGOLY \citep{Bean2024LINGOLYAB} leverage problem formats derived from linguistics olympiads, emphasizing deductive reasoning through tasks like unseen language translation, cognate identification, and graphophonemic transcription. While these tasks pose meaningful challenges, they tend to assess general reasoning rather than internalized linguistic knowledge relevant to natural language processing.

%  [Rephrase]
Our work diverges from these approaches by introducing problem sets that simultaneously assess linguistic knowledge and reasoning capabilities within the specific context of Korean language phenomena. Additionally, we establish empirical validation of our benchmark's ecological validity by demonstrating correlation between benchmark performance and human preference judgments—a methodological contribution absent in most previous linguistic evaluation frameworks.

\subsection{Korean Benchmarks} % 2.2 -----

% [Rephrase]
The landscape of Korean language benchmarks exhibits notable limitations in comprehensiveness, methodological rigor, and linguistic specificity. KLUE \citep{klue2021}, conceptualized as a Korean analog to GLUE, inherits the limitations of its predecessor by focusing on conventional NLP tasks that inadequately assess complex linguistic reasoning. Similarly, the Ko-H5 benchmark from Open Ko-LLM Leaderboard \citep{openkollm2024} relies primarily on translated English datasets, failing to capture language-specific linguistic phenomena unique to Korean.

% The landscape of Korean language benchmarks exhibits notable limitations in comprehensiveness, methodological rigor, and linguistic specificity. KLUE \citep{klue2021} focuses on simpler conventional NLP tasks, failing to assess diverse linguistic knowledge in depth. The Ko-H5 benchmark from the Open Ko-LLM Leaderboard \citep{openkollm2024} is based primarily on translated materials from English datasets. KoBEST \citep{jang-etal-2022-kobest} limits its coverage to word sense disambiguation and polarity. Hae-Rae bench \citep{son-etal-2024-hae} is based on crowd-sourced contents and existing materials such as questions from TV quiz shows, lacking expert knowledge. CLicK \citep{kim-etal-2024-click} presents unclear categorizations and relies simply on existing major examinations. KMMLU \citep{Son2024KMMLUMM} puts its focus on STEM disciplines rather than linguistic competence.
% [Rephrase]
KoBEST \citep{jang-etal-2022-kobest} represents a more linguistically-informed approach, comprising five tasks adapted from English benchmarks but designed by Korean linguists. However, its restricted scope—focusing primarily on word sense disambiguation and polarity classification—provides insufficient coverage of Korean's rich morphosyntactic, phonological, and pragmatic characteristics. This narrow assessment scope significantly limits its utility as a comprehensive measure of Korean linguistic competence.

% Recent benchmarks with broader coverage still encounter critical limitations. Hae-Rae Bench \citep{son-etal-2024-hae} emphasizes lexical knowledge and cultural understanding but relies heavily on crowdsourced content and existing materials such as quiz shows, introducing potential data contamination while lacking systematic linguistic framework. CLicK \citep{kim-etal-2024-click} presents an unclear categorization of linguistic phenomena and directly extracts questions from existing major examinations, raising serious concerns regarding both theoretical coherence and training data overlap. KMMLU \citep{Son2024KMMLUMM}, despite its extensive coverage across 45 subject areas, exhibits a pronounced emphasis on STEM disciplines at the expense of linguistic and cultural domains. Its methodology, based on automated extraction from Korean examinations, compounds data contamination risks while failing to systematically address rich Korean-specific characteristics.

% [Rephrase]
Recent benchmarks have attempted broader coverage but encounter methodological limitations. Hae-Rae Bench \citep{son-etal-2024-hae} emphasizes lexical knowledge and cultural understanding but relies heavily on crowdsourced content and existing quiz show materials, introducing potential data contamination while lacking systematic linguistic framework. CLicK \citep{kim-etal-2024-click} presents an ambiguous categorization of linguistic phenomena and directly extracts questions from standardized Korean examinations, raising serious concerns regarding both theoretical coherence and training data overlap.

% Together, these highlight three pervasive issues in existing Korean benchmarks: (1) over-reliance on translated or repurposed content rather than original materials; (2) insufficient engagement of linguistic expertise in benchmark construction; and (3) inadequate coverage of language-specific phenomena across core linguistic domains. Our benchmark addresses these limitations by involving native Korean linguists without relying on existing sources or machine translation. By systematically evaluating models across diverse linguistic domains through complex reasoning tasks, KoBALT provides a more robust assessment specifically designed to capture the diverse characteristics of the Korean language.

% [Rephrase]
These limitations collectively highlight three pervasive issues in existing Korean benchmarks: (1) overreliance on translated or repurposed content rather than originally developed materials; (2) insufficient engagement of linguistic expertise in benchmark construction; and (3) inadequate coverage of language-specific phenomena across core linguistic subfields. Our benchmark addresses these limitations through a comprehensive approach constructed entirely by native Korean linguists without reliance on existing sources or machine translation. By systematically evaluating models across diverse linguistic domains through complex reasoning tasks, KoBALT provides a more robust assessment framework specifically designed to capture the nuanced characteristics of Korean language.

\definecolor{syntax}{RGB}{25, 118, 210}     % 파란색
\definecolor{semantics}{RGB}{255, 111, 0}   % 밝은 주황색
\definecolor{pragmatics}{RGB}{0, 150, 136}  % 청록색
\definecolor{phonetics}{RGB}{156, 39, 176}  % 보라색
\definecolor{morphology}{RGB}{121, 134, 53} % 올리브 그린

\begin{figure}[htbp]
\centering
\begin{subfigure}[b]{0.45\textwidth}
    \centering
    \begin{tikzpicture}[scale=0.6]
        \pie[radius=1.8, text=pin, text=legend, sum=auto, before number=\footnotesize, after number=, font=\tiny, 
        color={syntax!95, syntax!80, syntax!65, syntax!50, syntax!35}]{
            28.7/Embedded Clauses (86),
            3.7/Ellipsis (11),
            34.7/Agreement (104),
            1.0/Scrambling (3),
            32.0/Arg. Structure (96)
        }
    \end{tikzpicture}
    \caption{\textbf{Syntax} (300)}
    \label{fig:pie_syntax}
\end{subfigure}
\hfill
\begin{subfigure}[b]{0.45\textwidth}
    \centering
    \begin{tikzpicture}[scale=0.6]
        \pie[radius=1.8, text=pin, text=legend, sum=auto, before number=\footnotesize, after number=, font=\tiny, 
        color={semantics!95, semantics!85, semantics!75, semantics!65, semantics!55, semantics!45, semantics!35}]{
            13.0/Ambiguity (27),
            13.0/Numeral Classifiers (27),
            11.5/Conjunctions (24),
            13.5/Rhetorical Expr. (28),
            13.5/Semantic Relations (28),
            28.8/Semantic Concord (60),
            10.1/Inter-sent. Relations (21)
        }
    \end{tikzpicture}
    \caption{\textbf{Semantics} (215)}
    \label{fig:pie_semantics}
\end{subfigure}

\vspace{0.3cm}
\begin{subfigure}[b]{0.45\textwidth}
    \centering
    \begin{tikzpicture}[scale=0.6]
        \pie[radius=1.8, text=pin, text=legend, sum=auto, before number=\footnotesize, after number=, font=\tiny, 
        color={pragmatics!95, pragmatics!80, pragmatics!65, pragmatics!50, pragmatics!35}]{
            27.2/Speech Acts (22),
            3.7/Relationship ID (3),
            27.2/Implicature (22),
            21.0/Deixis \& Reference (17),
            21.0/Conv. Principles (17)
        }
    \end{tikzpicture}
    \caption{\textbf{Pragmatics} (81)}
    \label{fig:pie_pragmatics}
\end{subfigure}
\hfill
\begin{subfigure}[b]{0.45\textwidth}
    \centering
    \begin{tikzpicture}[scale=0.6]
        \pie[radius=1.8, text=pin, text=legend, sum=auto, before number=\footnotesize, after number=, font=\tiny, 
        color={phonetics!95, phonetics!75, phonetics!55, phonetics!35}]{
            11.3/Basic Articul. Phonetics (7),
            54.8/Phonol. Alternation (34),
            22.6/Phonotactic Constr. (14),
            11.3/Suprasegmental (7)
        }
    \end{tikzpicture}
    \caption{\textbf{Phonetics \& Phonology} (62)}
    \label{fig:pie_phonetics}
\end{subfigure}

\vspace{0.3cm}
\begin{subfigure}[c]{0.32\textwidth}
    \centering
    \begin{tikzpicture}[scale=0.6]
        \pie[radius=1.8, text=pin, text=legend, sum=auto, before number=\footnotesize, after number=, font=\tiny, 
        color={morphology!95, morphology!75, morphology!55}]{
            19.0/POS \& Morphemes (8),
            28.6/Verbal Conjugation (12),
            52.4/Word Formation (22)
        }
    \end{tikzpicture}
    \caption{\textbf{Morphology} (42)}
    \label{fig:pie_morphology}
\end{subfigure}

\caption{Distribution of samples across linguistic domains in KoBALT}
\label{fig:pie_distributions}
\end{figure}
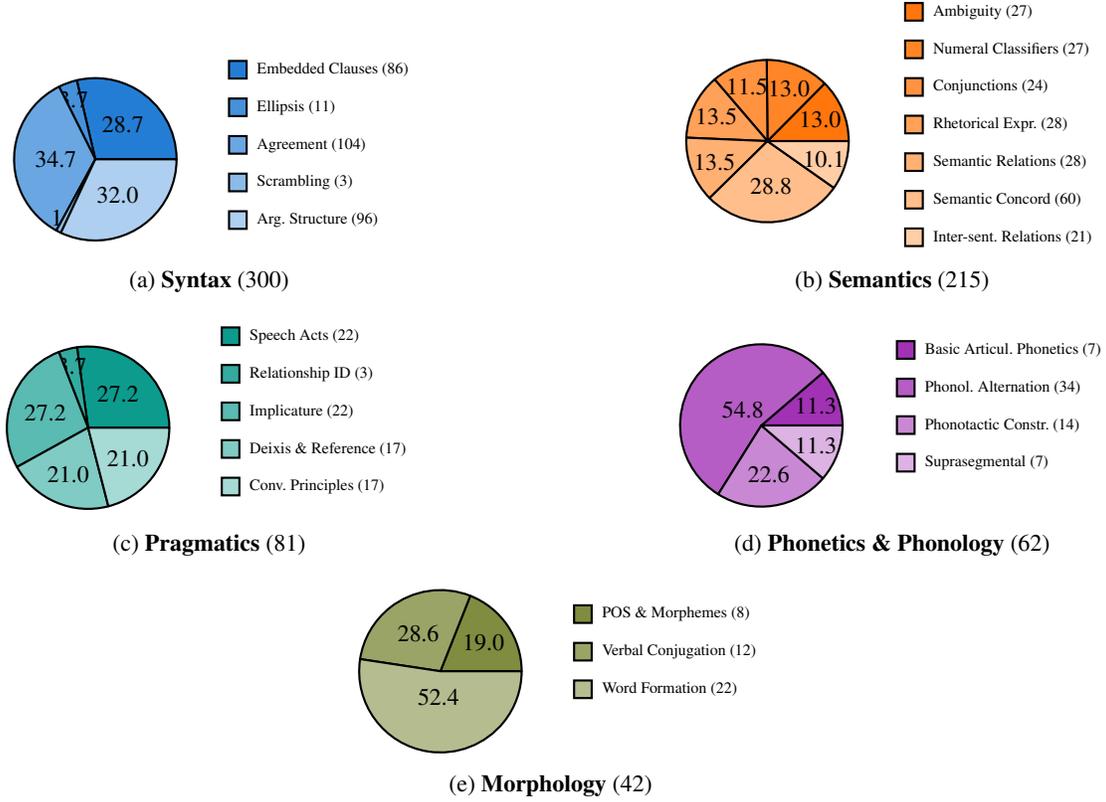

\section{KoBALT: Korean Benchmark for Advanced Linguistic Tasks}

% table 끝

KoBALT comprises 700 theoretically motivated, linguist-crafted multiple-choice questions designed to systematically assess language models' competence over 24 linguistic phenomena across five fundamental domains of linguistics, namely \textit{Syntax}, \textit{Semantics}, \textit{Pragmatics}, \textit{Phonetics and Phonology}, and \textit{Morphology}. Each question presents ten choices, challenging models to demonstrate both linguistic knowledge and reasoning capabilities. The benchmark was meticulously developed through a structured methodology that involved seven linguistically trained annotators who constructed, validated, and refined the set of questions according to rigorous criteria. The sample questions of our benchmark is illustrated in Table \ref{tab:kobalt_samples}

% 문항 샘플 테이블
\begin{table}[htbp]
\centering
\footnotesize
\begin{tabular}{|>{\centering\arraybackslash}p{1cm}|>{\centering\arraybackslash}p{2cm}|p{12cm}|}
\hline
\textbf{Domain} & \textbf{Ling. Phenomenon} & \textbf{Sample Question} \\
\hline
\centering Morpho-logy & \centering Part of Speech \& Morphemes & 
\begin{tabular}[c]{@{}p{12cm}@{}}
\textbf{문제:} 문장 내에서 '들'이 조사로서 사용되지 않은 경우는? \\
(Question: In which case is '들' not used as a particle in a sentence?) \\
\textbf{정답:} 너희'들' 빨래는 다 널었니? \\
(Answer: Did you[plural] hang all the laundry?) \\
\textbf{오답:} 어서'들' 오세요. 외 9개 \\
(Incorrect: Please[plural] come in quickly. plus 9 other options)
\end{tabular} \\
\hline
\centering Syntax & \centering Embedded Clauses & 
\begin{tabular}[c]{@{}p{12cm}@{}}
\textbf{문제:} 다음 문장은 서술어가 여러 개 있는 복합문이다. 이중 안긴 문장의 서술어의 자릿수가 1개인 것은? \\
(Question: The following sentences are complex sentences with multiple predicates. Which one has an embedded clause with a one-place predicate?) \\
\textbf{정답:} 고객님, 점선을 따라서 편안한 숙소로 가세요. \\
(Answer: Sir/Madam, please follow the dotted line to the comfortable accommodation.) \\
\textbf{오답:} 우리가 휴양지에 갔다는 사실을 말하지 말아 줘. 외 9개 \\
(Incorrect: Please don't tell anyone the fact that we went to the resort. plus 9 other options)
\end{tabular} \\
\hline
\centering Phonetics \& Phonology & \centering Phonological Alternation & 
\begin{tabular}[c]{@{}p{12cm}@{}}
\textbf{지문:} <보기> 값[갑] \\
(Text: <Example> 값[갑] - price/value [gap]) \\
\textbf{문제:} <보기>의 예시에서 나타난 음운 현상이 나타나지 않는 것을 고르시오. \\
(Question: Choose the word where the phonological phenomenon shown in the example does not occur.) \\
\textbf{정답:} 깎다 \\
(Answer: to cut/trim [kkakkda]) \\
\textbf{오답:} 외곬; 읊고; 값; 맑게; 넓다; 넋; 훑고; 밝지; 묽고 \\
(Incorrect: one-track [oegot]; reciting [eulpgo]; value [gap]; clearly [malge]; wide [neopda]; soul [neokk]; scanning [heultgo]; brightly [balkji]; thin [mulggo])
\end{tabular} \\
\hline
\centering Seman-tics & \centering Semantic Concord Between Predicates and Arguments & 
\begin{tabular}[c]{@{}p{12cm}@{}}
\textbf{지문:} 연휴도 끝나가는 시점이라, [회사1]로 출근한 동료는 거의 볼 수 없었다. [구내 식당]에도, [탕비실]에도, 또 평소엔 발 디딜 틈 하나 없던 [휴게실]에도 인적이 끊겨 한산하다. [집1]에서 [회사2]까지, 또 [회사3]에서 [집2]까지 이동하는 시간을 고려하면 밀린 일이 많지 않은 이상 합리적인 선택이라고 생각한다. \\
(Text: As the holiday was coming to an end, I could hardly see colleagues who came to [company1]. [Cafeteria], [pantry], and even [lounge] that is usually too crowded to step in, are all empty and quiet. Considering the commute time from [home1] to [company2], and from [company3] to [home2], I think it's a reasonable choice unless there's a lot of pending work.) \\
\textbf{문제:} 다음 지문의 괄호 중 의미적 역할이 가장 유사한 것들로 구성된 선지를 고르시오. \\
(Question: Choose the option that consists of bracketed terms with the most similar semantic roles.) \\
\textbf{정답:} [회사1], [탕비실], [휴게실] \\
(Answer: [company1], [pantry], [lounge]) \\
\textbf{오답:} [구내 식당], [탕비실], [집1] 외 9건 \\
(Incorrect: [cafeteria], [pantry], [home1] plus 9 other options)
\end{tabular} \\
\hline
\centering Prag-matics & \centering Implicature & 
\begin{tabular}[c]{@{}p{12cm}@{}}
\textbf{문제:} 갑의 질문에 대해 을의 대답을 통해 긍정/부정을 판단할 수 있는 예가 아닌 것은? \\
(Question: Which is NOT an example where you can determine affirmation/negation from B's answer to A's question?) \\
\textbf{정답:} 갑: 따님은 공부를 잘 하나요? / 을: 애가 머리는 저를 안 닮고 애엄마를 닮았어요. \\
(Answer: A: Does your daughter study well? / B: The child takes after her mother's intelligence, not mine.) \\
\textbf{오답:} 갑: 영화 좋아하세요? / 을: 전 활동적인 취미가 좋더라고요. 외 9건 \\
(Incorrect: A: Do you like movies? / B: I prefer more active hobbies. plus 9 other options)
\end{tabular} \\
\hline
\end{tabular}
\caption{Sample Questions from KoBALT}
\label{tab:kobalt_samples}
\end{table}

The data construction followed an expert-driven approach guided by five central methodological principles:

\begin{description}[leftmargin=0pt, style=unboxed]
\setlength{\itemsep}{0pt}
\item[Comprehensive Linguistic Coverage:] To establish a taxonomically valid representation of Korean linguistic phenomena, we conducted a systematic analysis of educational frameworks including the Standard Curriculum for Korean Language (both L1 and L2 variants) and high-stakes assessment instruments (Korean Language Test, Test Of Proficiency In Korean, College Scholastic Ability Test, and Public Service Examination). Analyzing these, we identified distinct linguistic phenomena often addressed in Korean language education, categorized them into topics in each linguistic domain, and supplemented them with additional phenomena characteristic of the Korean language based on theoretical consideration.
Through this process, we established a comprehensive taxonomic structure encompassing 24 distinct linguistic phenomena distributed across five domains. 
% The details are illustrated in Table \ref{fig_phenomena}. For a more detailed description of each phenomenon, see Appendix \ref{appendix_phenomena}.
\item[Structural and Lexical Diversity:] To mitigate potential confounds from question format bias, we implemented deliberate variation in question structures within each phenomenon category. While maintaining the basic 10-choice question format, we employed diverse question types to prevent model overfitting to specific question patterns. Our question sets include multiple strategies which require models to solve various tasks such as cloze-style completion, grammaticality judgment, minimal pair contrast, and context interpretation.

\item[Naturalness and Representativeness:] All materials were carefully crafted to reflect the linguistic intuition and real-world usage of the Korean language while maintaining precise control over target phenomena. In terms of lexical diversity, annotators incorporated vocabulary from multiple difficulty levels, referencing the vocabulary grading data from the National Institute of Korean Language (NIKL) to ensure balanced representation across all complexity tiers. \footnote{The data is licensed under Korea Open Government License (BY-NC-ND). (URL: \url{https://korean.go.kr/front/reportData/reportDataView.do?mn_id=207&report_seq=1160})} Further, all questions were cross-examined by multiple annotators to validate their grammaticality, naturalness, clarity, and accurate representation of the target phenomenon.
\item[Use of Linguistic Knowledge and World Knowledge:] Our question sets were designed to evaluate models based on their accurate knowledge of linguistic phenomena and proper reasoning skills rather than random guesses. For this, KoBALT utilizes a 10-choice format, which provides a broader range of distracting options, increasing the difficulty of the whole benchmark and lowering the probability of mistakenly overestimating the model's competence based on random guesses. Further, we intentionally include some questions requiring the models to utilize both their linguistic knowledge and world knowledge. This is inevitable for proper assessment of linguistic knowledge in some phenomena such as hypernymy and other semantic relations.
% For detailed descriptions, see Appendix \ref{appendix_A}.

\subsection{Assigning Difficulty Levels}
To provide a meaningful categorization of questions for better analysis, we classified our benchmark items into three difficulty levels in a post hoc manner based on the performance patterns of leading models, from the easiest (Level 1) to the hardest (Level 3). We selected four high-performing models as difficulty indicators: Claude-3.7-Sonnet, Claude-3.5-Sonnet, GPT-4o, and DeepSeek-V3. Questions were assigned to difficulty categories using the following criteria:

\begin{itemize}
\item Level 1 (Easy): All four selected models answered correctly
\item Level 2 (Intermediate): Two or three models answered correctly
\item Level 3 (Hard): Zero or one model answered correctly
\end{itemize}

%------ 난이도 테이블 (public 기준) -----
\begin{table}[h]
\centering
\begin{tblr}{
  hline{1,5} = {-}{0.08em},
  hline{2} = {-}{},
}
Level            & \# of Questions \\
1 (Easy)         & 182             \\
2 (Intermediate) & 220             \\
3 (Hard)         & 298             
\end{tblr}
\caption{Distribution of Questions in 3 Difficulty Levels}
\label{tab_difficulty}
\end{table}

This classification approach leverages the collective performance of diverse state-of-the-art models to establish a robust difficulty scale. Since these four models represent the current upper boundary of performance, this categorization should remain relevant for evaluating both current and future models. The resulting distribution of questions across difficulty levels is presented in Table \ref{tab_difficulty}.

\end{description}

% \usepackage{multirow}
% \usepackage{multirow}
% \usepackage{multirow}
% \usepackage{multirow}
% \usepackage{multirow}
% \usepackage{multirow}

\subsection{Syntax (300 Questions) }

Our syntax component represents the largest subset of the benchmark, covering five phenomena fundamental to Korean syntactic structure: \textit{Agreement}, \textit{Argument Structure and Valency}, \textit{Embedded Clauses}, \textit{Ellipsis}, and \textit{Scrambling}. The first three are selected based on grammar taught in Korean secondary education and constructions frequently featured in Korean language proficiency assessments. In contrast, Ellipsis and Scrambling are included to capture Korean-specific syntactic patterns, such as flexible word order and omitted constituents, which are not explicitly covered in the high school curriculum but are common in real-world usage.

\begin{description}[leftmargin=0pt, style=unboxed]
\setlength{\itemsep}{0pt}
\item[\textit{Agreement(Agr)}] focuses on syntactic dependencies between grammatical elements, including subject-verb agreement, honorific agreement, negative polarity item (NPI) licensing, tense–aspect alignment, and voice alternations involving passive and causative constructions. Korean exhibits a rich honorific system where verbal forms and particles reflect social relationships between discourse participants, resulting in agreement patterns that differ from those found in many Indo-European languages. Questions require models to detect agreement violations and infer their causes, group expressions that share the same type of agreement pattern, and actively manipulate sentence structure—such as through voice alternation—to preserve grammatical agreement.

\item[\textit{Argument Structure and Valency(ASV)}] examines knowledge of predicate-argument relationships, focusing on valency and case realization. In Korean, case interpretation is often complicated by the omission of case markers, the presence of auxiliary particles, and flexible word order. Questions assess models' ability to determine the valency of predicates and to identify the grammatical roles of constituents such as subjects, objects, and obligatory adjuncts.

\item[\textit{Embedded Clauses(EC)}] evaluates comprehension of complex clausal structures, primarily focusing on embedded clauses. Korean embedded clauses reflect typologically distinctive properties, including head-final structure and highly productive clause transformation strategies that allow embedded predicates to be realized in a wide range of syntactic functions. Questions require models to identify whether a sentence contains an embedded clause, determine the grammatical role of a constituent within the embedded clause, and classify the function of the embedded clause itself.

\item[\textit{Ellipsis(Elp)}] focuses on Korean-specific grammatical omission patterns. Korean permits extensive null arguments whose referents must be recovered from discourse context, posing significant challenges for computational models. Questions assess models’ ability to recover elided content, distinguish between grammatical and ungrammatical ellipsis, and determine whether omission alters the original meaning of the sentence..

\item[\textit{Scrambling(Scr)}] focuses on Korean's relatively free word order phenomena, where arguments can appear in multiple positions without altering grammatical relations. Questions require models to identify natural word orders, detect ungrammatical scrambling, and determine whether a given constituent can be felicitously inserted into specific syntactic positions.
\end{description}

\subsection{Semantics (215 Questions) }

Our semantics component addresses seven  phenomena central to meaning construction in Korean: \textit{Semantic Concord between Predicates and Arguments}, \textit{Rhetorical Expressions}, \textit{Ambiguity}, \textit{Semantic Relationships between Words}, \textit{Numeral Classifiers}, \textit{Conjunctions}, and \textit{Inter-sentence Semantic Relationship}. These categories reflect semantic features that are frequently tested  in Korean language proficiency assessments. In establishing this classification scheme, we consider both the classification widely used in linguistic theory and the step-by-step process by which lexical items combine to form sentence-level meaning.

\begin{description}[leftmargin=0pt, style=unboxed]
\setlength{\itemsep}{0pt}
\item[\textit{Semantic Concord between Predicates and Arguments(SCPA)}] examines thematic role assignment, and semantic concord between predicates and their arguments. Korean predicates impose specific semantic constraints on arguments, including animacy requirements, intentionality features, and scalar properties that must be satisfied for well-formedness. Questions assess models’ ability to detect predicate-argument incompatibility and assign correct thematic roles.

\item[\textit{Rhetorical Expressions(RE)}] targets non-literal language usage including metaphor, irony, hyperbole, and idiomatic expressions. Korean employs culturally-specific figurative expressions and conventionalized rhetorical patterns that require both linguistic and cultural knowledge for correct interpretation. Questions assess models’ ability to recognize figurative meaning and appropriately select rhetorical devices.

\item[\textit{Ambiguity(Amb)}] focuses on lexical, structural, and scope ambiguities in Korean. Questions require detecting sentence-level ambiguity and identifying multiple possible interpretations, as well as analyzing the source of ambiguity and proposing appropriate disambiguation strategies.

\item[\textit{Semantic Relationships between Words(SRW)}] assesses understanding of lexical semantic relations including synonymy, antonymy, hyponymy, meronymy, and semantic frames. Korean exhibits specialized taxonomic relationships encoded through compound formation. Questions test models’ ability to identify semantic relationships, recognize incompatible semantic features, and categorize lexical items according to shared semantic properties.

\item[\textit{Numeral Classifiers(NC)}] evaluates knowledge of Korean's extensive classifier system, where numeric expressions require specific classifier morphemes based on semantic properties of the quantified noun. Questions require models to select appropriate numeral classifiers for given referents within context.

\item[\textit{Conjunctions(Conj)}] examines semantic properties of coordinating and subordinating conjunctions that signal logical entailment, temporal order, and causal relations at the sentence-to-sentence level. Korean conjunctions encode fine-grained distinctions in temporal overlap, causality strength, and conditionality that must be precisely interpreted. Questions assess whether models can select conjunctions that naturally connect given sentences, infer appropriate continuations based on a provided sentence and conjunction, and judge whether the rhetorical effect of a conjunction is accurately described.

\item[\textit{Inter-sentence Semantic Relationship(ISR)}] focuses on the use of phrases, clauses, or sentences that contributes to maintaining coherence across a discourse. Questions evaluate models’ ability to select the most contextually appropriate unit (e.g., phrase, clause, or sentence) to complete a passage, as well as to perform higher-level operations such as sentence ordering, removal of semantically irrelevant content, or judgment of logical consistency within the overall discourse.
\end{description}

\subsection{Pragmatics (81 Questions)}

Our pragmatics component addresses five phenomena central to contextual language interpretation: \textit{Implicature}, \textit{Speech Acts}, \textit{Conversational Principles and Discourse Strategy}, \textit{Deixis and Reference}, and \textit{Inter-person Relationship}. Unlike other domains, our selection on pragmatic phenomena is based on the taxonomy of the field's academic research, as pragmatic competence is typically acquired through communicative experience and is rarely covered in explicit curricula or standardized assessments. 

We select \textit{Implicature}, \textit{Speech Acts}, and \textit{Deixis and Reference} as core phenomena widely discussed in pragmatic theory. To capture the models' discourse-level interpretation, we add \textit{Conversational Principles and Discourse Strategy}. Lastly, we include \textit{Inter-person Relationship} to reflect a distinctive feature of Korean, where the same sentential meaning can take different forms depending on interpersonal relationships.

\begin{description}[leftmargin=0pt, style=unboxed]
\setlength{\itemsep}{0pt}
\item[\textit{Implicature(Impl)}] assesses understanding of implied meanings beyond literal semantic content. Utterances involving conversational implicature often reflect cultural and conventional norms, and implicature in Korean likewise exhibits culture-specific characteristics. Questions require models to recover the implied meaning of certain utterance in a given conversation, while some ask to choose a proper utterance when given a surrounding context and the intended meaning.

\item[\textit{Speech Acts(SA)}] evaluates comprehension of speech acts, including statement, expressive, promise, question, directive/command and proposing. Korean employs grammatical markers, politeness forms, and a rich system of rhetorical questions, all of which play a crucial role in the interpretation of indirect speech acts and speaker intentions. Questions test models’ ability to identify speech act types indicating the intent of the speaker, recognize indirect speech act and pair certain utterances with same speech act types based on given utterances.

\item[\textit{Conversational Principles and Discourse Strategy(CPDS)}] examines models' knowledge on speakers’ various discourse strategies including adherence or violation of various conversational principles for certain intents. Questions assess recognition of intentional violation on certain conversational maxims, and interpretation of certain pragmatic and metapragmatic signals and their hidden intents behind them based on given conversations.

\item[\textit{Deixis and Reference(DR)}] focuses on context-dependent reference expressions, including personal, spatial, temporal, and social deixis. Korean employs multiple demonstrative systems, specialized social deictic markers, and complex referential expressions sensitive to social relations. Questions test models’ ability to resolve referential expressions, track referents across discourse, and interpret deictic terms relative to contextual parameters.

\item[\textit{Inter-person Relationship(IPR)}] assesses understanding of Korean's elaborate social indexing system, including honorifics, address terms, and speech level markers that reflect social relationships between discourse participants. Questions require models to infer social relationships from linguistic forms.
\end{description}

\subsection{Phonetics and Phonology (62 Questions)}

Our phonetics and phonology component covers four phenomena fundamental to Korean sound patterns: \textit{Phonological Alternation}, \textit{Phonological  Constraints}, \textit{Basic Articulatory Phonetics}, and \textit{Suprasegmental Features}. Among these, P\textit{honological Alternation}, \textit{Phonological Constraints}, and \textit{Suprasegmental Features} fall under phonology and are selected with reference to the Korean secondary school curriculum and Korean language proficiency assessments. In contrast, Basic Articulatory Phonetics corresponds to phonetics. Among phenomena addressed in phonetics, we include only articulatory aspects, as they are most compatible with text-based evaluation using language models.

\begin{description}[leftmargin=0pt, style=unboxed]
\setlength{\itemsep}{0pt}
\item[\textit{Phonological Alternation(PA)}] examines systematic sound changes including substitution, deletion, contraction, and insertion. Substitution includes fortition, syllable-final neutralization and assimilation. Deletion encompasses consonant cluster simplification and vowel deletion. Contraction involves aspiration and vowel contraction. Insertion covers /n/-insertion and the use of the interfix -s- (known as Sai-siot in Korean). Questions require models to understand the concept of phonological alternation and to determine whether a given word undergoes a specific type of alternation.

\item[\textit{Phonological Constraints(PC)}] focuses on permissible sound sequences in Korean, incorporating phonotactic, syllabic and word-level constraints. Questions require models to determine whether a word conforms to certain types of constraints and to assess the correctness of general descriptive statements about phonotactic patterns in Korean.

\item[\textit{Basic Articulatory Phonetics(AP)}] evaluates understanding of speech sound production. Questions cover major articulatory distinctions such as place of articulation (e.g., bilabial, dental, palatal, velar, glottal) and manner of articulation (e.g., stops, affricates, fricatives, nasals, laterals), as well as additional features such as aspiration (unaspirated vs. aspirated), continuancy (continuant vs. non-continuant), and sonority (sonorant vs. obstruent).

\item[\textit{Suprasegmental Features(SF)}] examines prosodic elements in Korean that crucially contributes to the meaning of an utterance, including vowel length distinctions (long vs. short), types of interrogatives, sentence-final endings, and intonation patterns. Questions assess models’ ability to identify vowel length or intonation contours and to determine the sentence type based on prosodic cues and sentence-final endings.
\end{description}

\subsection{Morphology (42 Questions)}

Our morphology component examines three phenomena fundamental to the internal structure of words of Korean: \textit{Word Formation}, \textit{Verbal Conjugation}, and \textit{Part-of-Speech and Morphemes}. These categories reflect frequently tested features of Korean morphology in Korean language proficiency assessments. Each category is designed to address a distinct and non-overlapping aspect of morphological knowledge—ranging from how words are formed, to how they are grammatically inflected, to how they are structurally categorized.

\begin{description}[leftmargin=0pt, style=unboxed]
\setlength{\itemsep}{0pt}
\item[\textit{Word Formation(WF)}] focuses on two core morphological processes: derivation and compounding. Korean exhibits rich morphological variation, requiring a thorough analysis of root forms and affix positions. Questions require models to recover original morphemes from surface forms, classify compound words by structural or semantic composition, and judge whether the resulting words are morphologically well-formed.

\item[\textit{Verbal Conjugation(VC)}] examines morphological patterns in the inflection of Korean verbs and adjectives, focusing on how stems and endings combine and change during conjugation. Korean exhibits both regular and irregular conjugation patterns, with irregular forms involving systematic but non-standard changes to verb stems and endings. Questions require models to recover base forms from inflected expressions, classify verbs and adjectives by conjugation pattern, and judge the morphological validity of derived forms based on stem-affix interactions.

\item[\textit{Part of Speech and Morphemes(POSM)}] addresses part-of-speech classification and morpheme-level analysis. Questions assess models’ ability to determine the part of speech of a given word and classify each morpheme into the appropriate morphological type.
\end{description}

% --------------- 4. Experiment ---
\section{Model Performances on KoBALT}

\subsection{Evaluation}
\label{sec_evaluation}

Using KoBALT, we evaluate the competence on Korean language of 20 LLMs across various model families including open-source and proprietary ones.

\begin{figure}[t]
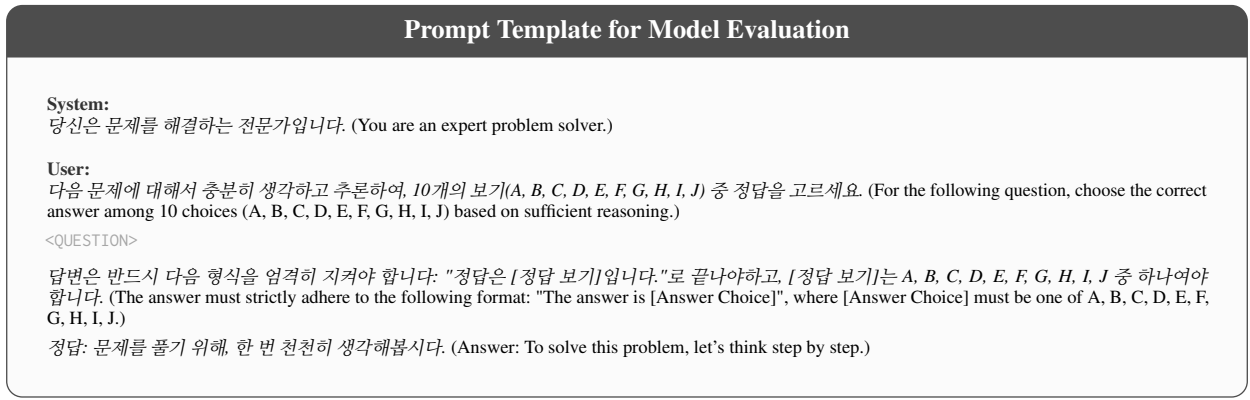

\centering
\begin{tcolorbox}[
    width=\linewidth,
    colback=gray!3,
    colframe=black!70,
    boxrule=0.5pt,
    arc=2mm,
    fontupper=\scriptsize,
    title={\centering\textbf{Prompt Template for Model Evaluation}},
    left=10pt,
    right=10pt,
    top=10pt,
    bottom=10pt,
    boxsep=5pt,
]
\begin{minipage}{\linewidth}
\textbf{\textcolor{black!80}{System:}}\\  
\textit{당신은 문제를 해결하는 전문가입니다.} (You are an expert problem solver.)
\vspace{3mm}

\textbf{\textcolor{black!80}{User:}}\\  
\textit{다음 문제에 대해서 충분히 생각하고 추론하여, 
10개의 보기(A, B, C, D, E, F, G, H, I, J) 중 정답을 고르세요.} (For the following question, choose the correct answer among 10 choices (A, B, C, D, E, F, G, H, I, J) based on sufficient reasoning.)
\vspace{1mm}

\textcolor{gray!70}{\texttt{<QUESTION>}}
\vspace{2mm}

\textit{답변은 반드시 다음 형식을 엄격히 지켜야 합니다: "정답은 [정답 보기]입니다."로 끝나야하고, 
[정답 보기]는 A, B, C, D, E, F, G, H, I, J 중 하나여야 합니다.} (The answer must strictly adhere to the following format: "The answer is [Answer Choice]", where [Answer Choice] must be one of A, B, C, D, E, F, G, H, I, J.)
\vspace{1mm}

\textit{정답: 문제를 풀기 위해, 한 번 천천히 생각해봅시다.} (Answer: To solve this problem, let's think step by step.)
\end{minipage}
\end{tcolorbox}
\caption{Example prompt used for model response generation. Models were instructed to select from ten possible choices after analyzing the linguistic task. The strict output format facilitated automated evaluation of model responses.}
\label{fig_eval_prompt}
\end{figure}

\begin{description}[leftmargin=0pt, style=unboxed]
    \setlength{\itemsep}{0pt}
    \item [Model] For open-source models, we select LLaMA 3.1 8B Instruct \citep{grattafiori2024llama3herdmodels}, LLaMA 3.3 70B Instruct\footnote{\url{https://huggingface.co/meta-llama/Llama-3.3-70B-Instruct}}, Mistral 7B Instruct \citep{jiang2023mistral7b}, Ministral 8B Instruct\footnote{\url{https://huggingface.co/mistralai/Ministral-8B-Instruct-2410}}, Mistral Small 3.1 24B Instruct \citep{mistral2024small31}, Gemma 2 9B Instruct \citep{gemmateam2024gemma2improvingopen}, Gemma 3 27B Instruct \citep{gemmateam2025gemma3technicalreport}, Qwen 2.5 7B Instruct, Qwen 2.5 32B Instruct, Qwen 2.5 72B Instruct \citep{Yang2024Qwen25TR}, Aya Expanse 8B, and Aya Expanse 32B \citep{dang2024ayaexpansecombiningresearch}.
    % Exaone 3.5 7.8B Instruct and Exaone 3.5 32B Instruct \citep{research2024exaone35serieslarge}. 
    For closed-source models, we evaluate GPT-4o \citep{gpt42023}, Claude 3.5 Sonnet \citep{anthropic2024claude35}, Claude 3.7 Sonnet \citep{anthropic2024claude37}, C4ai Command A 03 \citep{cohere2025commandaenterprisereadylarge}, DeepSeek V3\footnote{\texttt{DeepSeek-V3-0324}, evaluated using fp8 precision via Together API.} and DeepSeek V3 XL\footnote{\texttt{unsloth/DeepSeek-V3-0324-GGUF/UD-Q2\_K\_XL}, evaluated using 2.71-bit quantization via llama.cpp.} \citep{deepseekai2025deepseekv3technicalreport}.
    \item [Prompt]
    The prompt used for response generation is illustrated in Figure \ref{fig_eval_prompt}. Models are required to provide its answer after a reasoning step.
    \item [Implementation Detail] To ensure reproducibility, we control $\texttt{temperature=0.0}$ for proprietary models, and set $\texttt{do\_sample=False}$ for open-source models. We control response length by setting $\texttt{max\_new\_tokens=2048}$ in all cases. 3 of A100 GPUs are used in evaluating dataset with 70B-72B series models, 2 for 27B-32B models and one for the other small models. It took 8 hours for LLaMA 3.3 70B, and much shorter for smaller models.  
    \item [Metric]
    Since the prompt explicitly requires the models to provide their answer in the format \textit{The answer is [Answer Choice]}, we extract the [Answer Choice] portion using regular expressions and compare it with the gold label.  A prediction is considered correct if the extracted answer matches the gold label exactly, and incorrect otherwise. Accuracy—defined as the proportion of correctly predicted instances among the total number of questions—is reported as our primary evaluation metric.
    
\end{description}

% prompt example

% ---4.3 Main Result ----------------------
\begin{table}[t]
\centering
\small
\resizebox{\linewidth}{!}{%
\begin{tblr}{
  width = \linewidth,
  colspec = {Q[220]Q[80]Q[80]Q[80]Q[80]Q[80]},
  hline{1,22} = {-}{0.08em},
  hline{2} = {-}{},
}
Model & Average & Syntax & Semantics & Pragmatics & Morphology & Phonetics\\
GPT-4o & 0.44 & 0.45 & 0.55 & 0.40 & 0.17 & 0.26\\
C4ai-command-a-03 & 0.36 & 0.30 & 0.52 & 0.36 & 0.24 & 0.18\\
Claude-3-5-sonnet & 0.52 & 0.52 & 0.65 & 0.51 & 0.36 & 0.24\\
Claude-3-7-sonnet & \textbf{0.61} & \textbf{0.66} & \textbf{0.66} & \textbf{0.64} & \textbf{0.36} & \textbf{0.31}\\
DeepSeek-V3-XL & 0.47 & 0.49 & 0.56 & 0.42 & 0.24 & 0.29\\
DeepSeek-V3 & 0.43 & 0.41 & 0.57 & 0.42 & 0.26 & 0.23\\
\hline
% EXAONE-3.5-32B & 0.29 & 0.23 & 0.43 & 0.26 & 0.17 & 0.16\\
% EXAONE-3.5-7.8B & 0.25 & 0.19 & 0.40 & 0.20 & 0.17 & 0.13\\
Qwen2.5-32B & 0.30 & 0.23 & 0.49 & 0.28 & 0.21 & 0.11\\
Qwen2.5-72B & \textbf{0.37} & \textbf{0.33} & \textbf{0.51} &\textbf{ 0.37 }& \textbf{0.24} & \textbf{0.18}\\
Qwen2.5-7B & 0.19 & 0.14 & 0.33 & 0.11 & 0.19 & 0.06\\
Aya-expanse-32b & 0.25 & 0.21 & 0.40 & 0.12 & 0.10 & 0.16\\
Aya-expanse-8b & 0.19 & 0.15 & 0.33 & 0.11 & 0.12 & 0.06\\
Gemma-2-9b & 0.21 & 0.17 & 0.34 & 0.15 & 0.12 & 0.11\\
Gemma-3-27b & 0.35 & 0.30 & 0.53 & 0.27 & 0.24 & 0.11\\
Llama-3.1-8B & 0.17 & 0.13 & 0.26 & 0.12 & 0.10 & 0.11\\
Llama-3.3-70B & 0.32 & 0.25 & 0.50 & 0.35 & 0.17 & 0.15\\
Ministral-8B & 0.17 & 0.11 & 0.29 & 0.15 & 0.10 & 0.11\\
Mistral-7B-v0.3 & 0.12 & 0.11 & 0.16 & 0.11 & 0.14 & 0.06\\
Mistral-Small-24B & 0.32 & 0.27 & 0.49 & 0.30 & 0.21 & 0.11\\
\end{tblr}
}
\caption{Performance of language models across linguistic domains in KoBALT}
\label{table:model_performance}
\end{table}

\begin{figure}[h] % h: here, 이미지 위치 제어
    \centering
    \includegraphics[width=1.0\textwidth]{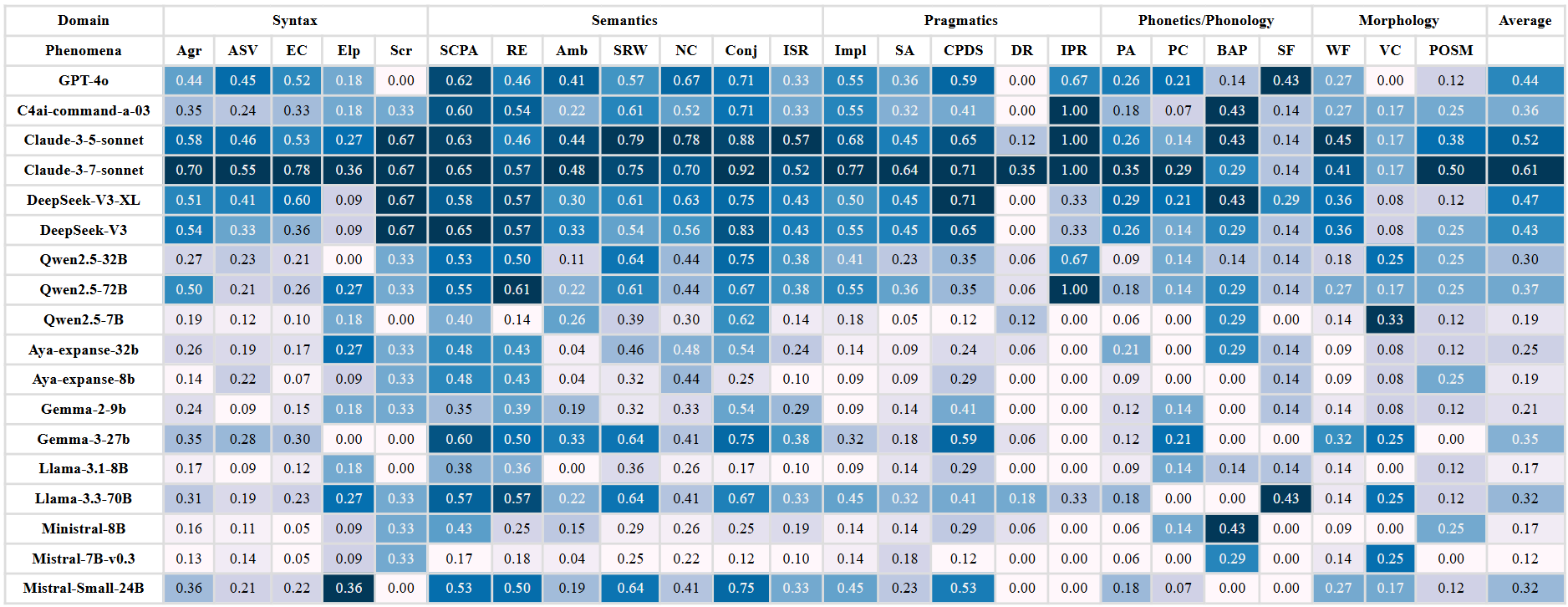} % 이미지 파일 이름 (확장자 없이도 가능)
    \caption{Accuracy per each linguistic phenomena}
    \label{fig:accuracy_per_phenomena}
\end{figure}

\textbf{General analysis.} The evaluation results on KoBALT are illustrated in Table \ref{table:model_performance}. The best performing model was Claude-3.7-Sonnet-20250219, which scored 0.61. Overall, the cutoff point emerges around 0.4 in the average score. All proprietary models and Deepseek-V3 exceed 0.4, with Deepseek-V3 (fp8) scoring 0.43 at the lowest. In contrast, all open-source models, except for Deepseek-V3, scored below 0.4, with Qwen2.5-72B scoring the highest at 0.37. The general trend observed is as follows: model performance improved with increasing model size (100B+ > 70B > 32B > 8B), but tended to decline as the granularity of linguistic unit in question increased, with the most prevalent pattern being: Semantics > Pragmatics > Syntax > Morphology > Phonetics/Phonology. For more detailed analysis in terms of each linguistic phenomenon, refer to Figure \ref{fig:accuracy_per_phenomena}

\textbf{Analysis by model scale.} In general, larger models outperform smaller models. In 14 out of 24 linguistic phenomena, average performance was ranked as follows: 8B < 32B < 70B < 100B+. Among the ten exceptions, seven phenomena still align with the general trend, with a single reversal between adjacent sizes: 72B < 32B in four cases and 100B+ < 70B in three cases. The remaining three outliers are associated with Korean-specific linguistic features (verbal conjugation in Morphology, ellipsis in Syntax) or areas that require information less accessible to text-based models (place/manner of articulation in Phonology).

\textbf{Analysis by linguistic domain} 
Examining model performances across linguistic domains reveals distinctive patterns. Most models demonstrated stronger capabilities in Semantics (with Claude 3.5 Sonnet achieving 65\% in average and 44\% at its lowest as seen in Table \ref{table:model_performance} and \ref{table:semantics_performance} respectively) but struggled significantly with Phonetics and Phonology tasks (where even top models fell below 45\%, as evidenced in Table \ref{table:phonetics_performance}).
% with Claude 3.5 Sonnet achieving 59.8\% as seen in Table \ref{table:semantics_performance 

This performance disparity challenges traditional linguistic assumptions, which typically associate higher levels of abstraction with increased processing complexity. As shown in Table \ref{table:syntax_performance}, model performances on Syntax (Claude 3.5 Sonnet: 52.6\%) were generally robust but still trailed Semantics. Similarly, results on Pragmatics in Table \ref{table:pragmatics_performance} reveal moderate performance (Claude 3.5 Sonnet: 51.0\%), while Morphology in Table \ref{table:morphology_performance} indicates models tend to face more struggles on the domain (with most models below 50\%).

One possible explanation for this pattern is that semantic, pragmatic, and syntactic processing may benefit from the basic training process of LLMs. Unlike humans, these models are trained via auto-regressive next-token prediction on massive text corpora, which naturally emphasizes contextual relationships between tokens. This training paradigm appears to effectively support higher-level abstract understanding required in these domains.

The pronounced difficulties in Phonetics/Phonology (Table \ref{table:phonetics_performance}) and Morphology (Table \ref{table:morphology_performance}) likely stem from multiple factors: these areas often require explicit linguistic rules that may be underrepresented in text-based training data; they frequently involve specialized terminologies; and they deal with fine-grained language units (sounds, morphemes) that may be less directly captured in text representations. Additionally, the text-centric nature of pretraining datasets might lead to uneven knowledge distribution across linguistic levels, with particular gaps in speech-related phenomena that might benefit from audio training data.

These analyses are supported by the observation that models tend to perform better in areas that closely align with characteristics of the training data. Certain linguistic phenomena, such as \textit{Implicature} and \textit{Conversational Principles and Discourse Strategy} in Pragmatics (Table \ref{table:pragmatics_performance}), as well as \textit{Word Formation} and \textit{Part-of-Speech and Morpheme}(Table \ref{table:morphology_performance}), appear particularly relevant to the types of data typically leveraged during training. Considering that large language models are trained through instruction tuning to capture user intentions and enhance conversational abilities, models’ aptitude in pragmatics is somewhat predictable. Likewise, traditional NLP tasks such as part-of-speech tagging are likely to contribute to models’ morpheme-level understanding by providing clues for morphological analysis. These tasks may assist models to identify a word’s grammatical category, grasp the relationship among its constituent morphemes, and derive word-internal structure. Based on the agglutinative nature of Korean, in which grammatical morphemes are affixed to convey additional meaning, one plausible generalization is that morphological patterns can be deduced when the root is retained after derivation or inflection, allowing models to trace word-internal structure.

In contrast, the models struggled with metalinguistic abstraction involving Korean-specific patterns, as reflected in the following areas: \textit{Ellipsis} in Syntax (< 55\%, in Table \ref{table:syntax_performance}), \textit{Ambiguity} in Semantics (<49\%, in Table \ref{table:semantics_performance}), \textit{Deixis and Reference} in Pragmatics(<36\%, in Table \ref{table:pragmatics_performance}), \textit{Verbal Conjugation} in Morphology (<34\%, in Table \ref{table:morphology_performance}), and \textit{Phonological Constraints} in Phonetics and Phonology (<30\%, in Table \ref{table:phonetics_performance}). This might stem from the distinct pattern of the language in use, such as frequent usage of Korean-specific lexical items like unique terms of address and reference, seen in \textit{Deixis and Reference}. Other Korean-specific patterns also seem to cause difficulties to the models, as seen in \textit{Verbal Conjugation} in Morphology where verbs displaying both regular and irregular conjugating patterns are tackled. With subword tokenization, models may struggle with forming a correct morphological paradigm that captures irregular variations and suppletions, which are distinct from base and regular word-forms and thus challenging to explain by systematic rule application on shared subword tokens.

%--- Syntax 테이블 ------
\begin{table}[h]
\centering
\small
\resizebox{0.9\linewidth}{!}{%
\begin{tblr}{
  width = \linewidth,
  colspec = {Q[220]Q[90]Q[90]Q[90]Q[70]Q[70]},
  hline{1,22} = {-}{0.08em},
  hline{2} = {-}{},
}
Model & Agreement & Arg. Structure & Embedded Clauses & Scrambling & Ellipsis\\
Aya-expanse-32b & 0.26 & 0.19 & 0.17 & 0.33 & 0.27\\
Aya-expanse-8b & 0.14 & 0.22 & 0.07 & 0.33 & 0.09\\
C4ai-command-a-03 & 0.35 & 0.24 & 0.33 & 0.33 & 0.18\\
% EXAONE-3.5-32B & 0.31 & 0.16 & 0.17 & 0.67 & 0.55\\
% EXAONE-3.5-7.8B & 0.23 & 0.17 & 0.17 & 0.00 & 0.09\\
Qwen2.5-32B & 0.27 & 0.23 & 0.21 & 0.33 & 0.00\\
Qwen2.5-72B & 0.50 & 0.21 & 0.26 & 0.33 & 0.27\\
Qwen2.5-7B & 0.19 & 0.13 & 0.10 & 0.00 & 0.18\\
Claude-3-5-sonnet & 0.58 & 0.46 & 0.53 & 0.67 & 0.27\\
Claude-3-7-sonnet & 0.70 & 0.55 & 0.78 & 0.67 & 0.36\\
DeepSeek-V3-XL & 0.51 & 0.41 & 0.60 & 0.67 & 0.09\\
DeepSeek-V3 & 0.54 & 0.33 & 0.36 & 0.67 & 0.09\\
Gemma-2-9b & 0.24 & 0.09 & 0.15 & 0.33 & 0.18\\
Gemma-3-27b & 0.35 & 0.28 & 0.30 & 0.00 & 0.00\\
GPT-4o & 0.44 & 0.45 & 0.52 & 0.00 & 0.18\\
Llama-3.1-8B & 0.17 & 0.09 & 0.12 & 0.00 & 0.18\\
Llama-3.3-70B & 0.31 & 0.19 & 0.23 & 0.33 & 0.27\\
Ministral-8B & 0.16 & 0.11 & 0.05 & 0.33 & 0.09\\
Mistral-7B-v0.3 & 0.13 & 0.14 & 0.05 & 0.33 & 0.09\\
Mistral-Small-24B & 0.36 & 0.21 & 0.22 & 0.00 & 0.36\\
\end{tblr}
}
\caption{Performance of language models on Syntax categories}
\label{table:syntax_performance}
\end{table}

% Semantics 결과 테이블 
\begin{table}[h]
\centering
\small
\resizebox{0.9\linewidth}{!}{%
\begin{tblr}{
  width = \linewidth,
  colspec = {Q[170]Q[70]Q[80]Q[90]Q[90]Q[90]Q[90]Q[70]},
  hline{1,22} = {-}{0.08em},
  hline{2} = {-}{},
}
Model & Ambiguity & Semantic Relations & Semantic Concord & Rhetorical Expr. & Numeral Classifiers & Inter-sent. Relations & Conjunctions\\
Aya-expanse-32b & 0.04 & 0.46 & 0.48 & 0.43 & 0.48 & 0.24 & 0.54\\
Aya-expanse-8b & 0.04 & 0.32 & 0.48 & 0.43 & 0.44 & 0.10 & 0.25\\
C4ai-command-a-03 & 0.22 & 0.61 & 0.60 & 0.54 & 0.52 & 0.33 & 0.71\\
% EXAONE-3.5-32B & 0.15 & 0.54 & 0.55 & 0.46 & 0.44 & 0.24 & 0.46\\
% EXAONE-3.5-7.8B & 0.11 & 0.50 & 0.50 & 0.54 & 0.30 & 0.19 & 0.54\\
Qwen2.5-32B & 0.11 & 0.64 & 0.53 & 0.50 & 0.44 & 0.38 & 0.75\\
Qwen2.5-72B & 0.22 & 0.61 & 0.55 & 0.61 & 0.44 & 0.38 & 0.67\\
Qwen2.5-7B & 0.26 & 0.39 & 0.40 & 0.14 & 0.30 & 0.14 & 0.63\\
Claude-3-5-sonnet & 0.44 & 0.79 & 0.63 & 0.46 & 0.78 & 0.57 & 0.88\\
Claude-3-7-sonnet & 0.48 & 0.75 & 0.65 & 0.57 & 0.70 & 0.52 & 0.92\\
DeepSeek-V3-XL & 0.30 & 0.61 & 0.58 & 0.57 & 0.63 & 0.43 & 0.75\\
DeepSeek-V3 & 0.33 & 0.54 & 0.65 & 0.57 & 0.56 & 0.43 & 0.83\\
Gemma-2-9b & 0.19 & 0.32 & 0.35 & 0.39 & 0.33 & 0.29 & 0.54\\
Gemma-3-27b & 0.33 & 0.64 & 0.60 & 0.50 & 0.41 & 0.38 & 0.75\\
GPT-4o & 0.41 & 0.57 & 0.62 & 0.46 & 0.67 & 0.33 & 0.71\\
Llama-3.1-8B & 0.00 & 0.36 & 0.38 & 0.36 & 0.26 & 0.10 & 0.17\\
Llama-3.3-70B & 0.22 & 0.64 & 0.57 & 0.57 & 0.41 & 0.33 & 0.67\\
Ministral-8B & 0.15 & 0.29 & 0.43 & 0.25 & 0.26 & 0.19 & 0.25\\
Mistral-7B-v0.3 & 0.04 & 0.25 & 0.17 & 0.18 & 0.22 & 0.10 & 0.13\\
Mistral-Small-24B & 0.19 & 0.64 & 0.53 & 0.50 & 0.41 & 0.33 & 0.75\\
\end{tblr}
}
\caption{Performance of language models on Semantics categories}
\label{table:semantics_performance}
\end{table}

% Pragmatics 결과 테이블 
\begin{table}[h]
\centering
\small
\resizebox{0.9\linewidth}{!}{%
\begin{tblr}{
  width = \linewidth,
  colspec = {Q[220]Q[80]Q[80]Q[120]Q[120]Q[80]},
  hline{1,22} = {-}{0.08em},
  hline{2} = {-}{},
}
Model & Implicature & Speech Acts & Conv. Principles & Relationship ID & Deixis\\
Aya-expanse-32b & 0.14 & 0.09 & 0.24 & 0.00 & 0.06\\
Aya-expanse-8b & 0.09 & 0.09 & 0.29 & 0.00 & 0.00\\
C4ai-command-a-03 & 0.55 & 0.32 & 0.41 & 1.00 & 0.00\\
% EXAONE-3.5-32B & 0.41 & 0.14 & 0.47 & 0.00 & 0.06\\
% EXAONE-3.5-7.8B & 0.23 & 0.23 & 0.29 & 0.33 & 0.00\\
Qwen2.5-32B & 0.41 & 0.23 & 0.35 & 0.67 & 0.06\\
Qwen2.5-72B & 0.55 & 0.36 & 0.35 & 1.00 & 0.06\\
Qwen2.5-7B & 0.18 & 0.05 & 0.12 & 0.00 & 0.12\\
Claude-3-5-sonnet & 0.68 & 0.45 & 0.65 & 1.00 & 0.12\\
Claude-3-7-sonnet & 0.77 & 0.64 & 0.71 & 1.00 & 0.35\\
DeepSeek-V3-XL & 0.50 & 0.45 & 0.71 & 0.33 & 0.00\\
DeepSeek-V3 & 0.55 & 0.45 & 0.65 & 0.33 & 0.00\\
Gemma-2-9b & 0.09 & 0.14 & 0.41 & 0.00 & 0.00\\
Gemma-3-27b & 0.32 & 0.18 & 0.59 & 0.00 & 0.06\\
GPT-4o & 0.55 & 0.36 & 0.59 & 0.67 & 0.00\\
Llama-3.1-8B & 0.09 & 0.14 & 0.29 & 0.00 & 0.00\\
Llama-3.3-70B & 0.45 & 0.32 & 0.41 & 0.33 & 0.18\\
Ministral-8B & 0.14 & 0.14 & 0.29 & 0.00 & 0.06\\
Mistral-7B-v0.3 & 0.14 & 0.18 & 0.12 & 0.00 & 0.00\\
Mistral-Small-24B & 0.45 & 0.23 & 0.53 & 0.00 & 0.00\\
\end{tblr}
}
\caption{Performance of language models on Pragmatics categories}
\label{table:pragmatics_performance}
\end{table}

% Phonology & Phonetics 결과 테이블 
\begin{table}[h]
\centering
\small
\resizebox{0.9\linewidth}{!}{%
\begin{tblr}{
  width = \linewidth,
  colspec = {Q[220]Q[120]Q[120]Q[120]Q[120]},
  hline{1,22} = {-}{0.08em},
  hline{2} = {-}{},
}
Model & Phonological Constr. & Phonol. Alternation & Suprasegmental & Basic Articul. Phonetics\\
Aya-expanse-32b & 0.00 & 0.21 & 0.14 & 0.29\\
Aya-expanse-8b & 0.00 & 0.09 & 0.14 & 0.00\\
C4ai-command-a-03 & 0.07 & 0.18 & 0.14 & 0.43\\
% EXAONE-3.5-32B & 0.00 & 0.21 & 0.29 & 0.14\\
% EXAONE-3.5-7.8B & 0.00 & 0.18 & 0.00 & 0.29\\
Qwen2.5-32B & 0.14 & 0.09 & 0.14 & 0.14\\
Qwen2.5-72B & 0.14 & 0.18 & 0.14 & 0.29\\
Qwen2.5-7B & 0.00 & 0.06 & 0.00 & 0.29\\
Claude-3-5-sonnet & 0.14 & 0.26 & 0.14 & 0.43\\
Claude-3-7-sonnet & 0.29 & 0.35 & 0.14 & 0.29\\
DeepSeek-V3-XL & 0.21 & 0.29 & 0.29 & 0.43\\
DeepSeek-V3 & 0.14 & 0.26 & 0.14 & 0.29\\
Gemma-2-9b & 0.14 & 0.12 & 0.14 & 0.00\\
Gemma-3-27b & 0.21 & 0.12 & 0.00 & 0.00\\
GPT-4o & 0.21 & 0.26 & 0.43 & 0.14\\
Llama-3.1-8B & 0.14 & 0.09 & 0.14 & 0.14\\
Llama-3.3-70B & 0.00 & 0.18 & 0.43 & 0.00\\
Ministral-8B & 0.14 & 0.06 & 0.00 & 0.43\\
Mistral-7B-v0.3 & 0.00 & 0.06 & 0.00 & 0.29\\
Mistral-Small-24B & 0.07 & 0.18 & 0.00 & 0.00\\
\end{tblr}
}
\caption{Performance of language models on Phonetics \& Phonology categories}
\label{table:phonetics_performance}
\end{table}

% Mophology 결과 테이블 
\begin{table}[h]
\centering
\small
\resizebox{0.9\linewidth}{!}{%
\begin{tblr}{
  width = \linewidth,
  colspec = {Q[220]Q[110]Q[110]Q[110]},
  hline{1,22} = {-}{0.08em},
  hline{2} = {-}{},
}
Model & Word Formation & Verbal Conjugation & POS \& Morphemes\\
Aya-expanse-32b & 0.09 & 0.08 & 0.13\\
Aya-expanse-8b & 0.09 & 0.08 & 0.25\\
C4ai-command-a-03 & 0.27 & 0.17 & 0.25\\
% EXAONE-3.5-32B & 0.09 & 0.25 & 0.25\\
% EXAONE-3.5-7.8B & 0.14 & 0.25 & 0.13\\
Qwen2.5-32B & 0.18 & 0.25 & 0.25\\
Qwen2.5-72B & 0.27 & 0.17 & 0.25\\
Qwen2.5-7B & 0.14 & 0.33 & 0.13\\
Claude-3-5-sonnet & 0.45 & 0.17 & 0.38\\
Claude-3-7-sonnet & 0.41 & 0.17 & 0.50\\
DeepSeek-V3-XL & 0.36 & 0.08 & 0.13\\
DeepSeek-V3 & 0.36 & 0.08 & 0.25\\
Gemma-2-9b & 0.14 & 0.08 & 0.13\\
Gemma-3-27b & 0.32 & 0.25 & 0.00\\
GPT-4o & 0.27 & 0.00 & 0.13\\
Llama-3.1-8B & 0.14 & 0.00 & 0.13\\
Llama-3.3-70B & 0.14 & 0.25 & 0.13\\
Ministral-8B & 0.09 & 0.00 & 0.25\\
Mistral-7B-v0.3 & 0.14 & 0.25 & 0.00\\
Mistral-Small-24B & 0.27 & 0.17 & 0.13\\
\end{tblr}
}
\caption{Performance of language models on Morphology categories}
\label{table:morphology_performance}
\end{table}

\section{Do Results on KoBALT Align with Human Preferences?}
\subsection{Human Preference Evaluation}

\subsection{Result}
\begin{figure}[h] % h: here, 이미지 위치 제어
    \centering
    \includegraphics[width=\textwidth]{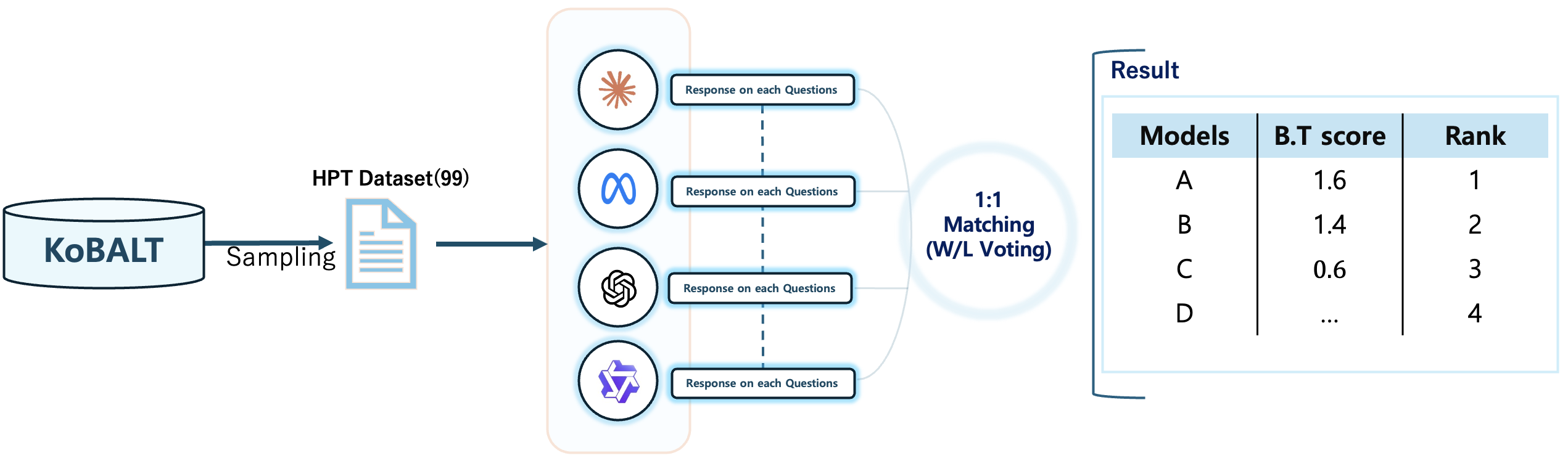} % 이미지 파일 이름 (확장자 없이도 가능)
    \caption{Human preference evaluation process utilizing the Online Live Chatbot Area concept}
    \label{fig:hpt_figure}
\end{figure}

As our practical goal lies in investigating if the models with stronger knowledge correlate with how native speakers perceive the models, we conducted a human preference evaluation. We adopt Chatbot Arena style evaluation \cite{chiang2024chatbotarenaopenplatform} for a carefully selected subset of our dataset. The overall process of this stage is illustrated in Figure \ref{fig:hpt_figure}.

From KoBALT, we sampled 99 problems representing each linguistic phenomenon. We selected 3 questions per each topic, considering the difficulty levels. We additionally selected 1-2 questions from some categories containing relatively more questions: \textit{Phonological Alternation} (Phonetics and Phonology), \textit{Semantic Concord between Argument and Predicates} (Semantics), \textit{Agreement}, \textit{Argument Structure and Valency}, and \textit{Embedded Clauses} (Syntax). For some phenomena which lacked questions in some difficulty levels, we sampled one from adjacent levels.

The questions were then rephrased into human-like queries to control the effect of unnecessary factors other than accuracy and factuality, such as response length \citep{dissectingfactor2024}, number of facts \citep{oh2024uncoveringfactorlevelpreferences}, formatting, and repetition \citep{hosking2024humanfeedbackgoldstandard}.

We leveraged responses from four large language models: two proprietary (Claude-3.5-Sonnet and GPT-4o) and two open-source (Qwen-2.5-72B and LLaMA-3.3-70B). Models were selected based on representativeness, performance, and parameter size. We gathered 95 participants (54 linguistics majors, 41 non-linguistics majors) who provided their preference votes over response pairs. 

\section{Interface and Instruction Used in Human Preference Collection}
\label{appendix_hpt_interface}

\begin{figure}[h] % h: here, 이미지 위치 제어
    \centering
    \includegraphics[width=0.8\textwidth]{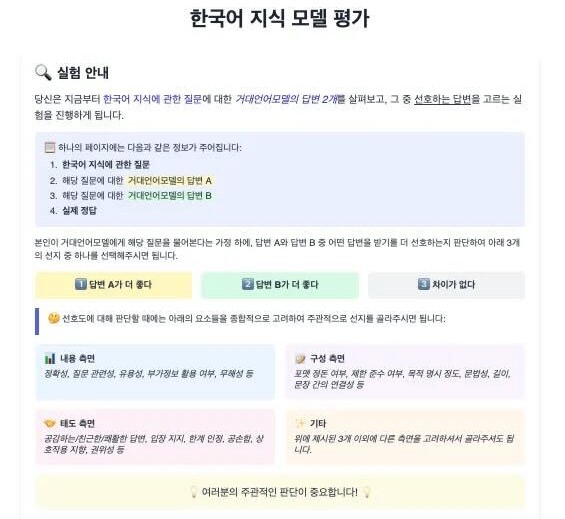}
    \caption{Screenshot of the full instruction given to participants in human preference collection stage.}
    \label{fig:hpt_screenshot}
\end{figure}

The full instruction given to human participants is illustrated in Figure \ref{fig:hpt_screenshot}, whose English translation is given in Figure \ref{appendix_hpt_interface_translation}:

\begin{figure}[t]
\centering
\begin{tcolorbox}[
    width=\linewidth,
    colback=gray!3,
    colframe=black!70,
    boxrule=0.5pt,
    arc=2mm,
    fontupper=\scriptsize,
    title={\centering\textbf{Full Instruction of Human Preference Test}},
    left=10pt,
    right=10pt,
    top=10pt,
    bottom=10pt,
    boxsep=5pt,
]
\begin{minipage}{\linewidth}
\textit{\textbf{Model Evaluation on Korean Linguistic Knowledge}}
\vspace{3mm}

\textbf{\textcolor{black!80}{User:}}\\  
\textit{Given the following two answers from LLMs about Korean language, you should choose one answer you prefer.}
\textit{Information provided are as follows:
\begin{enumerate}
    \item A question on Korean language
    \item Response from model A
    \item Response from model B
    \item Correct answer choice
\end{enumerate}}
\vspace{1mm}

\textit{Suppose you are the user who is asking the question, and decide which answer you would prefer. Choose one option from the following three choices.}
\textit{\begin{enumerate}
    \item Response A is better
    \item Response B is better
    \item There is no difference between the two responses.
\end{enumerate}}
\vspace{1mm}

\textit{Please consider various factors comprehensively, such as:}
\textit{\begin{itemize}
    \item Content: accuracy, relevance, usefulness, use of additional information, harmlessness
    \item Composition: formatting, adherence to the query, clear statement of its goal, grammaticality, length, cohesiveness
    \item Tone: Relating/Friendly/Lively tone,  stance, admittance of its limitation, politeness, openness, authoritativeness
    \item Others: You can make your decision according to other elements you regard as significant
\end{itemize}}
\vspace{1mm}

\textit{Please keep in mind that your subjective view matters to us.}
\vspace{1mm}

\end{minipage}
\end{tcolorbox}
\caption{Example prompt used for model response generation. Models were instructed to select from ten possible choices after analyzing the linguistic task. The strict output format facilitated automated evaluation of model responses.}

\label{appendix_hpt_interface_translation}
\end{figure}

\subsection{Results on Human Preference Evaluation}

\paragraph{Bradley--Terry Model}
To aggregate 51\,528 pairwise votes, we employed the
Bradley--Terry (BT) model.
For two systems $i$ and $j$, the win probability is
$\Pr(i \!\gg\! j)=\beta_i/(\beta_i+\beta_j)$.
Iterative maximum likelihood yields a \textit{normalized
BT score $\hat\beta$} for each system.
% \FloatBarrier

\paragraph{Overall and Domain-level Rankings}
The result is analyzed by the Bradley--Terry model, which produces a comparative ranking from pairwise games as mentioned before.
Table~\ref{tab:bt} reports the aggregate Bradley–Terry scores: 
\textbf{Claude-3.5-Sonnet} leads with \(\hat\beta=1.650\), followed by 
\textbf{GPT-4o} (1.018), 
\textbf{Qwen-2.5-72B} (0.878), 
and \textbf{Llama-3.3-70B} (0.514).\footnote{Ninety-five annotators produced 51\,528 comparisons; a “no difference’’ vote awarded 0.5 point to each side.} 
Overall, we confirm that the models’ quantitative performance (accuracy on KoBALT) closely mirrors human evaluators’ qualitative preferences.  
Table~\ref{tab:bt} breaks these scores down by five major linguistic domains. Generally, the rankings appeared similar to the overall rank (Claude-3.5-Sonnet > GPT-4o > Qwen-2.5-72B > LLaMA-3.3-70B), but with two notable exceptions. In Semantics, Qwen-2.5-72B (0.991) surpasses GPT-4o (0.840), and a similar downward shift in preference appears in Pragmatics domains where all models had excelled on the original benchmark. This pattern is especially pronounced for GPT-4o, indicating that high benchmark scores do not always lead to top qualitative impressions.
% \FloatBarrier

% ---------- TABLE 1 : BT scores by group ----------
\begin{table}[!htbp]
\centering\small
\begin{tabular}{lrrrr}
\toprule
\textbf{Domain} & \textbf{Claude} & \textbf{GPT-4o} & \textbf{Qwen} & \textbf{Llama} \\
\midrule
Overall                  & 1.650 & 1.018 & 0.878 & 0.514 \\
\cmidrule(lr){1-5}
Syntax                   & 1.710 & 1.072 & 0.780 & 0.514 \\
Semantics                & 1.681 & 0.840 & 0.991 & 0.559 \\
Pragmatics               & 1.769 & 0.998 & 0.808 & 0.506 \\
Phonetics/Phonology      & 1.520 & 1.121 & 0.962 & 0.441 \\
Morphology               & 1.368 & 1.285 & 0.888 & 0.492 \\
\bottomrule
\end{tabular}
\caption{Overall Bradley--Terry scores by linguistic domain(\textuparrow\ better).}
\label{tab:bt}
\end{table}

% ---------- TABLE 2 : Correlation table ----------
\begin{table}[!htbp]
\centering
\begin{tabular}{lrrr}
\toprule
\textbf{Model} & \textbf{Correlation} & \textbf{$p$-value} & $n$ \\
\midrule
Claude-3.5-Sonnet & 0.638 & 0.001 & 24 \\
GPT-4o            & 0.632 & 0.001 & 24 \\
Llama-3.3-70B     & 0.570 & 0.004 & 24 \\
Qwen-2.5-72B      & 0.570 & 0.004 & 24 \\
\bottomrule
\end{tabular}
\caption{Pearson correlation between phenomenon-level accuracy and BT score.}
\label{tab:corr-sub}
\end{table}

\paragraph{Correlation with KoBALT accuracy}
Phenomenon–level accuracy correlates positively with BT scores:
$r=0.638$ (Claude-3.5-Sonnet), $0.632$ (GPT-4o),
and $0.570$ for both Qwen-2.5-72B and Llama-3.3-70B
(*p* < .01; Table \ref{tab:corr-sub}).
Thus higher benchmark accuracy generally predicts stronger
human preference, although the gap between GPT-4o and Qwen remains modest. 

\paragraph{Pairwise Preference by Response Correctness}
We compare BT scores when both models are correct (\(O\) and \(O\)) versus when both are incorrect (\(X\) and \(X\)). As shown in Table~\ref{tab:bt_correct_wrong}, Claude-3.5-Sonnet is preferred most strongly in both settings (BT = 1.650 if both correct; BT = 1.414 if both wrong). Interestingly, GPT-4o and LLaMA-3.3-70B improve their relative standing on the “both wrong” condition, suggesting that their errors are perceived as less confusing than those of their peers.
% \FloatBarrier

% ---------- TABLE 3 : O X score table ----------
\begin{table}[!htbp]
\centering\small
\begin{tabular}{lrrr}
\toprule
Rank & Model               & \(O\)\&\(O\) & \(X\)\&\(X\) \\
\midrule
0    & Claude-3.5-Sonnet   & 1.650        & 1.414        \\
1    & GPT-4o              & 0.718        & 1.046        \\
2    & Qwen-2.5-72B        & 1.339        & 1.036        \\
3    & LLaMA-3.3-70B       & 0.266        & 0.582        \\
\bottomrule
\end{tabular}
\caption{BT scores when both models are correct (\(O\)\&\(O\)) or both incorrect (\(X\)\&\(X\)).}
\label{tab:bt_correct_wrong}
\end{table}

\paragraph{One-Correct Scenario}
When exactly one model’s answer is correct, we compute each model’s win rate. Table~\ref{tab:win_rate_one_correct} shows that Qwen-2.5-72B leads at 92.1\%, followed by Claude-3.5-Sonnet at 91.5\%. GPT-4o and LLaMA-3.3-70B record 87.7\% and 82.0\%, respectively, indicating that Qwen’s correct responses are the most persuasive.
% \FloatBarrier

% ---------- TABLE 4 : O X preference ratio table ----------
\begin{table}[htbp]
\centering\small
\begin{tabular}{lrr}
\toprule
Rank & Model               & Win\_Rate (\%) \\
\midrule
0    & Qwen-2.5-72B        & 92.1           \\
1    & Claude-3.5-Sonnet   & 91.5           \\
2    & GPT-4o              & 87.7           \\
3    & LLaMA-3.3-70B       & 82.0           \\
\bottomrule
\end{tabular}
\caption{Win rates when only one model’s answer is correct.}
\label{tab:win_rate_one_correct}
\end{table}

\paragraph{Qualitative Observations}
Through a qualitative analysis, we discovered three major findings. First, although Claude leads on every major phenomenon, \emph{Ambiguity} occasionally reverse the order of GPT-4o and Qwen. Second, preference gaps enlarge on easier items: when accuracy nears a ceiling, raters seem to rely on secondary cues — answer length,
politeness, style — making features unrelated to the response accuracy more decisive. Understanding which stylistic properties sway human judgement in this saturation regime needs to be dealth with in future study. Finally, Even after eliminating its ten malformed outputs,
Qwen remains third, indicating high persuasive power when its answers are well-formed.

% --- 7. 결론
\section{Discussion and Conclusion}
\label{conclusion}
We presented KoBALT, a benchmark of 700 linguist-crafted questions across 24 linguistic phenomena in Korean, designed to evaluate LLMs' linguistic competence with minimal training data overlap. Our evaluation of 20 LLMs showed that even the best-performing model (Claude 3.7 Sonnet) achieved only 61\% accuracy, with performance generally declining from semantics to phonetics/phonology across all models. Human preference evaluation with 95 annotators demonstrated significant correlation between benchmark scores and human judgments ($r=0.638$ for top models), validating KoBALT's effectiveness as a measure of Korean linguistic competence. The difficulty-stratified question set provides a framework for analyzing both current and future models. Our work addresses the need for linguistically-motivated evaluation in typologically diverse languages.
The main contributions of this work are as follows:

\begin{itemize}
\item We introduce \textbf{KoBALT}, a comprehensive Korean benchmark dataset for assessing advanced linguistic knowledge across five domains, featuring high-quality linguist-crafted questions targeting 24 distinct linguistic phenomena.
\item We provide systematic evaluation of 20 LLMs on KoBALT, revealing significant limitations in Korean linguistic competence even among state-of-the-art models and identifying patterns of strength and weakness across linguistic domains.
\item We establish the ecological validity of our benchmark through human preference evaluation, demonstrating strong alignment between KoBALT scores and Korean native speakers' judgments of model performance.
\end{itemize}

%-----------------------------

\section*{Limitations}
\label{limitations}
As our dataset is constructed in Korean, our dataset has limitations in evaluating models in other languages, especially English. Thus, the accuracy result and its correlation with human preference demonstrated in our paper are not guaranteed to be appropriate for other unilingual models. Moreover, our level system is based on model performance. This is based on our intuition that the powerful proprietary and open-source models could be the proper indicators to diagnose the difficulties of the questions. However, in further research, it should be conducted to establish a reliable level system in terms of humans and models. Lastly, in the human preference evaluation, we only considered 4 models, two for open-source and two for proprietary ones, considering the workload of the participants in the test. These are few considering the number of models being used in the quantitative accuracy result. The experiment with an enlarged setting should be conducted in further research.
\section{Ethics Statement}
In the data construction process, we carefully inspected all materials to not include any harmful or biased statements, real-world names, privacy invasions, or statements invoking any possible harassment over certain ethnic or religious groups. Our dataset is openly available and free of harmful content.
The Human Preference Evaluation in this study was conducted based on the Chatbot Arena-style evaluation. This method determines the ranking of model responses through anonymous votes collected via an online platform. Due to uncertainty within our institution regarding whether this method falls under IRB review requirements, we conducted an internal ethics review prior to proceeding with the preference evaluation in this format. Currently, several online platforms, including Huggingface, are available for evaluating human preferences for LLMs.

%\section*{Acknowledgements}

\bibliographystyle{unsrtnat}
\bibliography{references}  %%% Uncomment this line and comment out the ``thebibliography'' section below to use the external .bib file (using bibtex) .

\end{document}